\documentclass[a4paper,fleqn]{cas-dc}
\usepackage[numbers,sort&compress]{natbib}
\usepackage{algorithm}
\usepackage{threeparttable}
\usepackage{arydshln}
\usepackage{ulem}
\usepackage{adjustbox}
\usepackage{tabularx}
\usepackage{microtype}
\usepackage{float}
\usepackage{algorithm}
\usepackage{algpseudocode}

\ExplSyntaxOn
\keys_set:nn { stm / mktitle } { nologo }
\ExplSyntaxOff

\begin{document}
\let\WriteBookmarks\relax
\def\floatpagepagefraction{1}
\def\textpagefraction{.001}
\shorttitle{}
\shortauthors{Su et~al.}
\title [mode = title]{Harnessing Large Language Models for Biomedical Named Entity Recognition}

\author[1]{Jian Chen}
\credit{Investigation, Methodology, Software, Writing - original draft}

\author[2]{Leilei Su}
\credit{Writing - original draft, Writing - review \& editing}

\author[3]{Cong Sun}
\cormark[1]
\ead{csun.nlp@gmail.com}
\credit{Conceptualization, Writing - review \& editing, Supervision}

\address[1]{Department of Data Science and Big Data Technology, Hainan University, Haikou 570228, China}
\address[2]{Department of Mathematics, Hainan University, Haikou 570228, China}
\address[3]{Department of Population Health Sciences, Weill Cornell Medicine, New York 10022, USA}

\cortext[cor1]{Corresponding author}

\begin{abstract}
Background and Objective:
\newline
Biomedical Named Entity Recognition (BioNER) is a foundational task in medical informatics, crucial for downstream applications like drug discovery and clinical trial matching. However, adapting general-domain Large Language Models (LLMs) to this task is often hampered by their lack of domain-specific knowledge and the performance degradation caused by low-quality training data. To address these challenges, we introduce BioSelectTune, a highly efficient, data-centric framework for fine-tuning LLMs that prioritizes data quality over quantity. 
\newline
\newline
Methods and Results:
\newline
BioSelectTune reformulates BioNER as a structured JSON generation task and leverages our novel Hybrid Superfiltering strategy, a weak-to-strong data curation method that uses a homologous weak model to distill a compact, high-impact training dataset. 
\newline
\newline
Conclusions:
\newline
Through extensive experiments, we demonstrate that BioSelectTune achieves state-of-the-art (SOTA) performance across multiple BioNER benchmarks. Notably, our model, trained on only 50\% of the curated positive data, not only surpasses the fully-trained baseline but also outperforms powerful domain-specialized models like BioMedBERT.
\end{abstract}


\maketitle

\section{INTRODUCTION}

Large Language Models (LLMs), such as GPT-4 \cite{gpt-4}, have sparked a paradigm shift in Natural Language Processing (NLP), demonstrating exceptional performance across a wide spectrum of tasks. Pre-trained on vast text corpora, LLMs possess powerful generalization capabilities, enabling them to tackle complex problems through zero-shot and few-shot prompting \cite{few-shotlearners}. This has accelerated their adoption in diverse fields, including education, law, and healthcare.

In the biomedical domain, specialized LLMs like Med-PaLM2 \cite{Med-PaLM2}, PMC-Llama \cite{wu2024pmc}, and Chat-Doctor \cite{yunxiang2023chatdoctor} have shown promise in conversational and question-answering tasks. However, a significant performance gap remains when applying these models to fundamental information extraction tasks, particularly Biomedical Named Entity Recognition (BioNER). General-domain LLMs often lack the deep, domain-specific knowledge required to interpret complex biomedical texts accurately. Furthermore, studies have shown that generative LLMs tend to yield low precision and recall on NER tasks, failing to meet the high-accuracy demands of biomedical research \cite{chen2023robust, ye2023comprehensive}. As BioNER is a cornerstone for downstream applications such as drug discovery, gene function analysis, and clinical trial matching, bridging this performance gap is of critical importance.

To address these challenges, we formulate BioNER as an instruction-driven, structured data generation task. As illustrated in Figure 1, the model is provided with a piece of biomedical text and a specific instruction and is trained to generate a standardized, machine-readable JSON list of the identified entities. This approach not only unifies the extraction paradigm across different entity types but also capitalizes on the powerful instruction-following and text generation capabilities of modern LLMs. And we propose a novel framework to efficiently adapt general-domain LLMs for high-performance BioNER. Rather than relying on costly domain-specific pre-training, we focus on unlocking the potential of existing models through instruction tuning. We select the Qwen3 family of models as our foundation \cite{yang2025qwen3} and, to this end, curate and unify four benchmark BioNER datasets \cite{blurb} into an instruction-following format. The core of our framework is a novel data curation strategy we term "Hybrid Superfiltering," \cite{li2024superfiltering} which leverages a computationally inexpensive "weak" model to intelligently identify and select the most informative and difficult training samples for fine-tuning a more powerful "strong" model. Our main contributions are as follows:

\begin{itemize}

 \item We introduce Hybrid Superfiltering, a weak-to-strong data filtering strategy tailored for BioNER instruction tuning. By separating positive and negative samples and using a homologous weak model to score Instruction-Following Difficulty (IFD), this method curates a high-quality training subset that significantly boosts learning efficiency and model performance.

 \item We reformulate BioNER as an end-to-end text-to-structured-data generation task. By fine-tuning the LLM to directly output entities in a JSON format, we bypass the complexities of traditional sequence labeling and provide a clean, effective paradigm for solving information extraction tasks with generative models.

 \item Through extensive experiments, we demonstrate that our model, BioSelectTune trained on approximately 50\% of the highest-quality curated data, achieves state-of-the-art results on multiple in-domain and out-of-domain BioNER datasets. Our method surpasses not only powerful generalist models like GPT-4 but also domain-specialized models like BioMedBERT, validating its effectiveness and generalization capabilities.

\end{itemize}

\begin{figure*}
\centering
\includegraphics[width=0.9\textwidth]{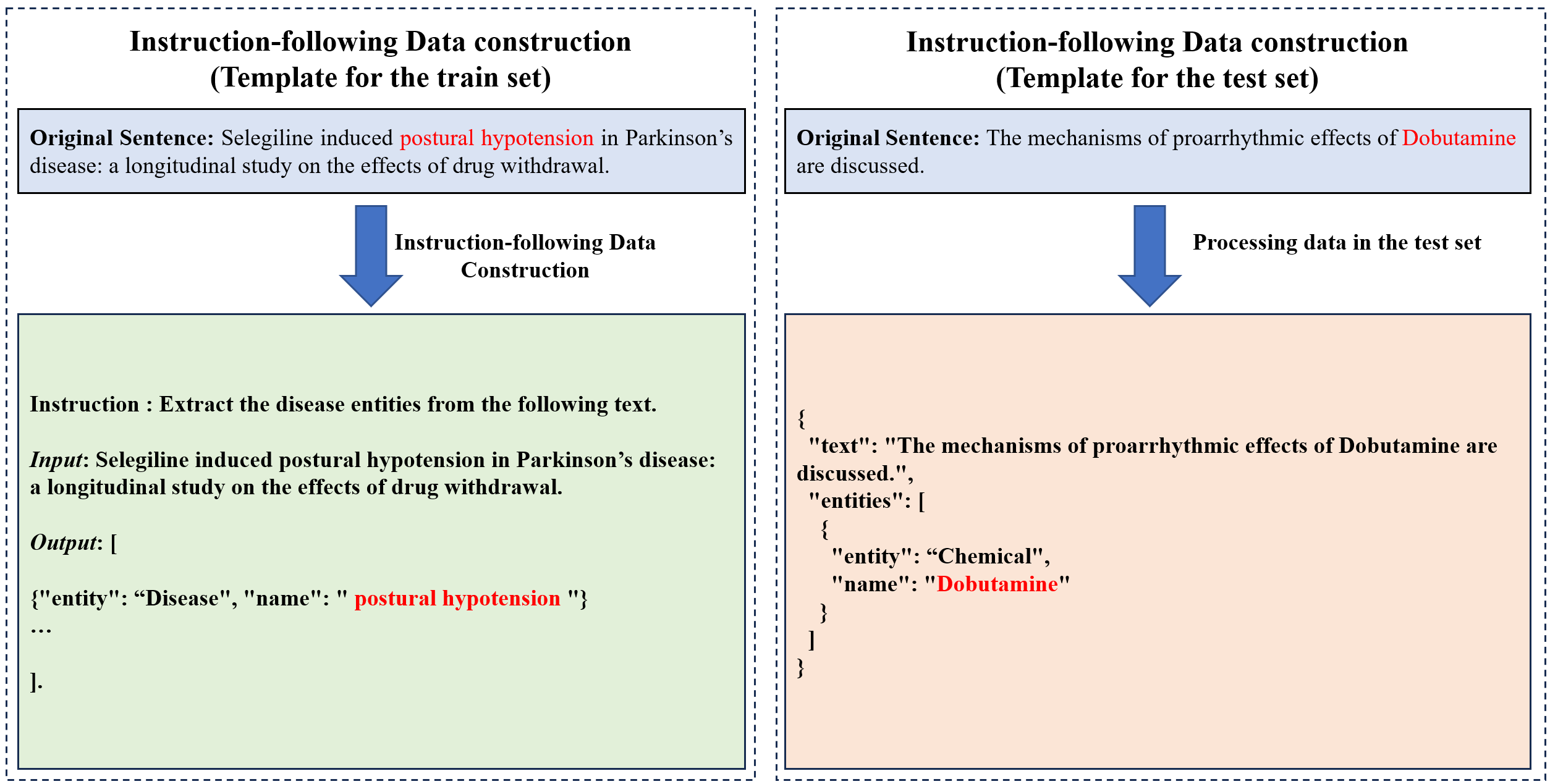}
\caption{Templates for instruction-following data and test data.}
\end{figure*}

\section{Related Work}

\textbf{LLMs for NER:}
Recent studies have explored the application of LLMs in NER by leveraging their generative nature. A notable example is GPT-NER, which adapts LLMs by converting sequence labeling into a generation task. This transformation uses specialized prompts and tokens, enhancing the performance of NER in various domains \cite{gpt-ner}. llmNER is a Python library for zero-shot and few-shot NER with LLMs, featuring an intuitive interface for prompt design, model querying, and output parsing, validated on multiple NER tasks \cite{villena2024llmner}. UniversalNER \cite{universalner} is a lightweight yet powerful model distilled from ChatGPT using mission-focused instruction tuning. It outperforms general instruction-tuned models like Alpaca \cite{alpaca}, Vicuna \cite{vicuna2023}, and InstructUIE \cite{instructuie} across domains, despite using far fewer parameters. In the biomedical domain, research has focused on enhancing in-context learning and fine-tuning strategies to improve LLM performance in BioNER tasks \cite{monajatipoor2024llms,taiyi}.

\textbf{Instruction Tuning:}
Instruction-tuning enables LLMs to follow natural language instructions for various downstream tasks. Given the high training costs of LLMs, this low-cost and efficient tuning method has attracted significant attention from researchers. For example, Tk-INSTRUCT, trained on a large-scale benchmark of diverse NLP tasks using instruction tuning, demonstrates strong generalization to unseen tasks, providing valuable insights for developing more general-purpose models \cite{tk}. Ouyang et al. collected high-quality instruction data using crowd-sourcing and fine-tuning GPT-3 to create InstructGPT, significantly improving its ability to understand user intent and follow instructions \cite{ouyang}. Recent developments \cite{alpaca,vicuna2023,peng2023instruction} have also led to smaller models that exhibit task-following capabilities after being fine-tuned on instruction data generated by LLMs like ChatGPT or GPT-4.

But the prevailing paradigms have largely overlooked a crucial question: can we achieve state-of-the-art performance not through more extensive pre-training or larger datasets, but through a more intelligent, data-centric approach to instruction tuning? This question is the primary motivation for our work. We hypothesize that the key to unlocking the full potential of general-purpose LLMs for specialized tasks like BioNER lies not in data quantity, but in its curated quality. We bridge this gap by proposing a novel framework that prioritizes the selection of a compact, high-impact training subset, aiming to match or even exceed the performance of domain-specialized models in a more efficient manner.


\section{Methodology}

Our proposed methodology advances the state-of-the-art in BioNER by synergizing an efficient data curation strategy with a structured generation paradigm. We introduce a novel Hybrid Superfiltering approach that leverages a weak language model to curate a highly informative, balanced dataset for instruction-tuning a powerful, larger model. The entire task is formulated as a unified structured data generation problem, where the model learns to output entities in a consistent JSON format. Our framework consists of three main stages: (1) formulating BioNER as a structured generation task, (2) curating the training data via our Hybrid Superfiltering strategy, and (3) fine-tuning the large language model on the curated data, as illustrated in Figure 2.

\begin{figure*}
\centering
\includegraphics[width=0.9\textwidth]{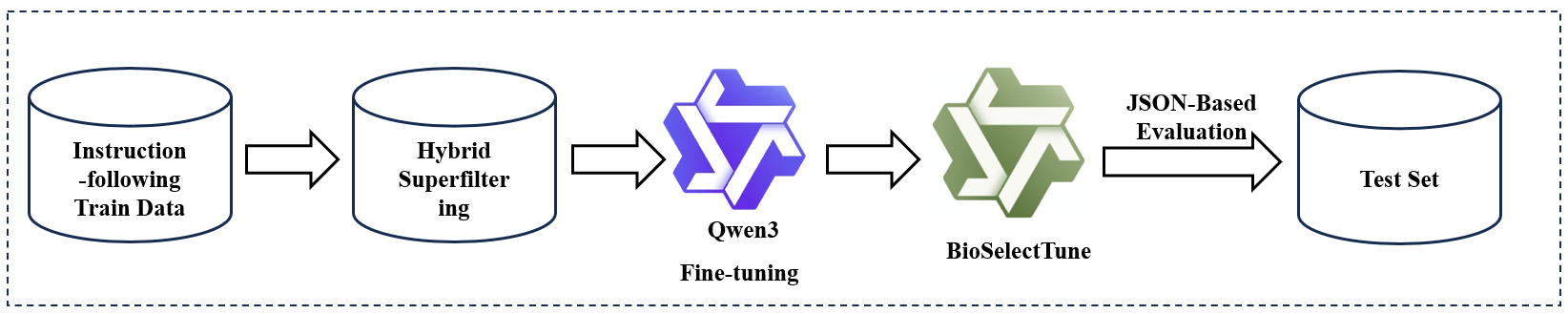}
\caption{Overall architecture of our method.}
\end{figure*}

\subsection{BioNER as a Structured Generation Task}

Traditional BioNER is often treated as a sequence labeling task. More recent approaches have reformulated it as a text-to-text generation task, using in-text markers to identify entities \cite{chen2025multimodal}. 
We depart from these paradigms and formulate BioNER as a text-to-structured-data generation task.

Given a raw dataset $\mathcal{D}_{\text{raw}} = \{(T_i, Y_i)\}_{i=1}^{N}$, where $T_i$ is a biomedical text and $Y_i$ is the set of ground-truth entities, we transform each instance into an instruction-following format. Each training example is a tuple $(x_i, y_i)$, where:

\begin{itemize}
    \item \textbf{Input ($x_i$):} The input is a concatenation of a task-specific instruction $I_i$ and the source text $T_i$, denoted as $x_i = I_i \oplus T_i$. For example, $I_i$ could be ``Extract the chemical entities from the following text.''.
    \item \textbf{Output ($y_i$):} The output $y_i$ is a structured representation of the entity set $Y_i$, serialized into a JSON array string. Each entity $e_j \in Y_i$ is represented as a key-value object with its type and name. Formally, $Y_i = \{e_j\}_{j=1}^{k_i}$, where $e_j = (c_j, n_j)$ consists of a category $c_j$ and a name $n_j$. The target output is its string representation:
    \begin{equation}
    y_i = \text{JSON}\left( \left\{ (\text{"entity"}, c_j), (\text{"name"}, n_j) \right\}_{j=1}^{k_i} \right)
\end{equation}
    
    Crucially, if a text $T_i$ contains no entities ($Y_i = \emptyset$, thus $k_i=0$), the target output is a string representing an empty array: $y_i = \text{``[]''}$.
\end{itemize}

The overall learning objective is to train a model with parameters $\theta$ to maximize the conditional probability of generating the correct structured output string $y_i$ given the input $x_i$, i.e., to learn the distribution $P_\theta(y|x)$.

\subsection{Hybrid Superfiltering}

A key challenge in instruction tuning is the presence of low-quality or redundant data. To address this, we adapt the ``Superfiltering'' \cite{li2024superfiltering} framework, a data curation strategy based on the weak-to-strong principle. Our novel approach, termed Hybrid Superfiltering, is specifically tailored for NER tasks by uniquely handling positive and negative samples.

\subsubsection{Homologous Weak-to-Strong Filtering via IFD Score}

The core premise of Superfiltering is that a computationally inexpensive, weaker model ($\mathcal{M}_{\text{weak}}$) can serve as an effective proxy to evaluate data quality for a stronger model ($\mathcal{M}_{\text{strong}}$) \cite{li2024superfiltering}. 
To maximize the fidelity of this proxy, we introduce a homologous filtering setup. Specifically, we select a weak model from the same family as our strong target model. In our experiments, we use Qwen3-0.6B as $\mathcal{M}_{\text{weak}}$ to curate data for Qwen3-8B ($\mathcal{M}_{\text{strong}}$). The rationale is that models from the same family often share architectural designs, tokenizers, and pre-training data distributions. This inherent alignment is hypothesized to yield a stronger correlation in their perception of instruction difficulty, making the weak model a more reliable and accurate proxy for the strong one.

We leverage the Instruction-Following Difficulty (IFD) score as the primary metric for data quality. The IFD score quantifies how much an instruction $x_i$ aids the model in generating the corresponding response $y_i$. It is defined as the ratio of the model's perplexity in generating $y_i$ with and without the instructional context $x_i$ \cite{li2024superfiltering}.
First, we define the perplexity (PPL) \cite{jelinek1977perplexity} of a sequence $S = (s_1, \ldots, s_L)$ given a model $\mathcal{M}$ as:

\begin{equation}
 \text{PPL}(S; \mathcal{M}) = \exp\left(-\frac{1}{L} \sum_{j=1}^{L} \log P_{\mathcal{M}}(s_j | s_{<j})\right) 
\end{equation}

The IFD score for a data pair $(x_i, y_i)$ is then calculated using our weak homologous model $\mathcal{M}_{\text{weak}}$:

\begin{equation}
 \text{IFD}(x_i, y_i) = \frac{\text{PPL}(y_i | x_i; \mathcal{M}_{\text{weak}})}{\text{PPL}(y_i; \mathcal{M}_{\text{weak}})} 
\end{equation}

where $\text{PPL}(y_i | x_i; \mathcal{M}_{\text{weak}})$ is the perplexity of generating the output conditioned on the input, and $\text{PPL}(y_i; \mathcal{M}_{\text{weak}})$ is the unconditional perplexity. A higher IFD score suggests that the instruction provides less direct guidance, indicating a more complex or informative sample.

\subsubsection{Hybrid Composition of Positive and Negative Samples}

A direct application of high-IFD filtering to our dataset revealed a critical insight: this method exclusively selects instances containing entities (positive samples) while discarding all instances without entities (negative samples, where $y_i = \text{``[]''}$). This is because generating an empty array is a simple task, resulting in a low IFD score.

However, negative samples are indispensable for robust NER performance, as they teach the model to avoid generating false positives. To resolve this, we devised a Hybrid Data Composition Strategy. Let the full instruction-formatted dataset be $\mathcal{D}$. We first partition it into positive and negative sets:

\begin{align}
     \mathcal{D}_{\text{pos}} = \{(x_i, y_i) \in \mathcal{D} \mid y_i \neq \text{``[]''}\} \\
     \mathcal{D}_{\text{neg}} = \{(x_i, y_i) \in \mathcal{D} \mid y_i = \text{``[]''}\} 
\end{align}

We then apply IFD-based filtering only to the positive set $\mathcal{D}_{\text{pos}}$. Following the methodology in \cite{li2024superfiltering}, we first discard samples with an IFD score greater than or equal to 1. From the remaining samples, we select the top-$\rho$ percentile based on the highest IFD scores to form a high-quality positive subset $\mathcal{D}'_{\text{pos}}$. The final training dataset $\mathcal{D}_{\text{train}}$ is constructed by combining this elite positive subset with the \textit{entire} set of negative samples:

\begin{equation}
    \mathcal{D}_{\text{train}} = \mathcal{D}'_{\text{pos}} \cup \mathcal{D}_{\text{neg}} 
\end{equation}

This hybrid approach ensures that the model is trained on the most informative entity-bearing examples while retaining comprehensive knowledge of non-entity contexts, striking an optimal balance between learning efficiency and model robustness.

\subsection{Instruction Fine-Tuning of the Generative Model}

The final stage involves fine-tuning the strong language model $\mathcal{M}_{\text{strong}}$ on the curated dataset $\mathcal{D}_{\text{train}}$. The model is trained to minimize the negative log-likelihood of the target JSON strings. The optimization objective is to find the optimal parameters $\theta^*$ that minimize the following loss function $\mathcal{L}(\theta)$:

\begin{equation}
     \mathcal{L}(\theta) = -\sum_{(x_i, y_i) \in \mathcal{D}_{\text{train}}} \sum_{j=1}^{|y_i|} \log P_{\theta}(y_{i,j} | y_{i, <j}, x_i) 
\end{equation}

where $|y_i|$ is the tokenized length of the target JSON string $y_i$, and $y_{i,j}$ is the $j$-th token. This standard auto-regressive training objective effectively teaches the model to generate well-formed JSON structures corresponding to the entities present in the input text, or an empty array otherwise.

\begin{algorithm}[H]
\caption{Hybrid Superfiltering and Fine-Tuning Framework}
\label{alg:framework}
\begin{algorithmic}[1]
\Require Full dataset $\mathcal{D}$, weak model $\mathcal{M}_{\text{weak}}$ (Qwen3-0.6B), strong model $\mathcal{M}_{\text{strong}}$ (Qwen3-8B), selection ratio $\rho$
\State Partition $\mathcal{D}$ into $\mathcal{D}_{\text{pos}}$ and $\mathcal{D}_{\text{neg}}$
\State Initialize an empty list $\mathcal{S}_{\text{IFD}}$
\ForAll{$(x_i, y_i) \in \mathcal{D}_{\text{pos}}$}
    \State Calculate $\text{IFD}(x_i, y_i)$ using Equation (3) with $\mathcal{M}_{\text{weak}}$
    \State Append $(\text{IFD}(x_i, y_i), (x_i, y_i))$ to $\mathcal{S}_{\text{IFD}}$
\EndFor
\State Sort $\mathcal{S}_{\text{IFD}}$ in descending order based on IFD scores
\State Let $k = \lfloor \rho \cdot |\mathcal{D}_{\text{pos}}| \rfloor$
\State Extract the top $k$ samples from the sorted list to form $\mathcal{D}'_{\text{pos}}$
\State Construct the final training set: $\mathcal{D}_{\text{train}} \leftarrow \mathcal{D}'_{\text{pos}} \cup \mathcal{D}_{\text{neg}}$
\State Fine-tune $\mathcal{M}_{\text{strong}}$ on $\mathcal{D}_{\text{train}}$ by minimizing the loss in Equation (7)
\State \textbf{return} Fine-tuned model $\mathcal{M}_{\text{strong}}$
\end{algorithmic}
\end{algorithm}

\section{Experiments}

\begin{table*}[htbp]
\centering
\caption{Model performance on in-domain biomedical datasets. Bolded values indicate the best performance. A model with the suffix `-SFT' denotes a fine-tuned version. The best F1-scores in each column are highlighted in bold.}
\scalebox{0.95}{
\begin{tabular}{lcccccccc}
\toprule
\textbf{Models} & \multicolumn{4}{c}{\textbf{F1-Score with Strict Match (\%)}} &  \\
& NCBI-Disease  & BC5CDR-Chemical  & BC5CDR-Disease  & BC2GM  \\
\midrule
GPT-4 & 67.40 & 83.70 & 67.80 & 57.20 \\
Taiyi & 73.10 & 80.20 & 69.10 & /  \\
InstructUIE-11B & 86.21 & / & / & \textbf{85.16} \\
UniNER-7B & 86.96 & 88.82 & 80.09 & 82.42\\
BioNER-Llama2-7B & 88.00 & 92.80 & 66.90 & 83.40 \\
BioMedBERT & 87.82 & \textbf{93.33} & 85.62 & 84.52 \\
\midrule
Qwen3-8B-SFT & 86.78 & 91.58 & \textbf{86.11} & 81.01  \\
BioSelectTune-8B (50\%) & \textbf{88.29} & 91.93 & 85.71 & 81.44  \\
\bottomrule
\end{tabular}
}
\end{table*}

\subsection{Experimental Datasets}

To rigorously evaluate our proposed methodology, we curated a comprehensive suite of benchmark datasets, partitioned into in-domain and out-of-domain categories. This selection facilitates a thorough assessment of both the model's primary task proficiency and its generalization capabilities on unseen data.

\subsubsection{In-domain Datasets}

The core of our fine-tuning and evaluation process utilized four widely recognized biomedical NER corpora. These datasets represent canonical entity types within the biomedical domain and serve to benchmark our model's performance on the primary tasks it was trained for. The selected datasets are:

\begin{itemize}
    \item \textbf{NCBI-Disease:} This corpus is a standard benchmark for disease name recognition, consisting of manually annotated PubMed abstracts \cite{dougan2014ncbi}.
    \item \textbf{BC5CDR (Chemical \& Disease):} The BioCreative V Chemical Disease Relation corpus contains PubMed abstracts annotated for both chemical and disease entities \cite{bc5cdr}. We treat the chemical and disease annotations as two distinct tasks for evaluation purposes (BC5CDR-Chemical and BC5CDR-Disease).
    \item \textbf{BC2GM:} The BioCreative II Gene Mention corpus is focused on the recognition of gene and gene product mentions in MEDLINE sentences, providing a benchmark for another critical entity type \cite{bc2gm}.
\end{itemize}
The inclusion of these datasets ensures a multifaceted evaluation of the model's ability to identify fundamental biomedical entities.

\subsubsection{Out-of-domain Datasets}

To assess the model's generalization performance and robustness, we conducted evaluations on two datasets that were not included in any stage of the fine-tuning process. These out-of-domain corpora allow us to measure the model's ability to adapt its learned knowledge to novel data distributions and annotation guidelines. The selected datasets are:

\begin{itemize}
    \item \textbf{NLM-Chem:} This dataset, from the BioCreative VII challenge, consists of full-text articles annotated for chemical entities, presenting a more complex challenge than abstract-only corpora \cite{islamaj2022nlm-chem}.
    \item \textbf{NLM-Gene:} This corpus provides annotations for gene mentions across multiple species within PubMed abstracts, testing the model's ability to handle ambiguous and diverse gene names \cite{islamaj2021nlm-gene}.
\end{itemize}
Performance on these datasets provides critical insights into the real-world applicability and generalization capacity of our fine-tuned model.

\subsection{Experimental Settings}

We compared our method with the following baselines: 
\begin{itemize}
\item GPT-4 \cite{gpt-4,SU2024104739}, a large language model developed by OpenAI. As a baseline for various information extraction tasks, GPT-4 demonstrates strong performance by leveraging its extensive pre-training on diverse textual data, enabling it to generate contextually relevant and coherent outputs even for complex queries.
\item InstructUIE \cite{instructuie}, a universal end-to-end information extraction framework that leverages natural language instructions to guide large models in performing IE tasks. 
\item UniversalNER \cite{universalner}, which proposed a more targeted distillation method and performed task-centered prompt fine-tuning, improving NER performance in most domains. 
\item Taiyi \cite{taiyi}, a bilingual (Chinese and English) large model fine-tuning on a substantial amount of biomedical data. 
\item BioNER-Llama2 \cite{keloth2024advancing}, a large language model built on Llama2-7B and designed for biomedical Named Entity Recognition. It utilizes a novel instruction-tuning paradigm to transform the NER task into a generative one, achieving performance comparable to specialized, fine-tuned biomedical models. 
\item BioMedBERT \cite{biomedbert}, a model pre-trained on a massive corpus of biomedical literature and clinical records, and one of the largest pre-trained models in the biomedical domain.    
\end{itemize}

All experiments were conducted based on the Qwen3-8B \cite{yang2025qwen3} model. We employed the Low-Rank Adaptation (LoRA) technique for parameter-efficient fine-tuning \cite{hu2022lora}. The model was trained for 3.0 epochs with a maximum sequence length of 128 tokens. We utilized a cosine learning rate scheduler with a peak learning rate of $1.0 \times 10^{-4}$ and accelerated training with BF16 mixed-precision. For evaluation, model performance was measured by the micro-F1. 

\subsection{Main Results}

\subsubsection{Performance on In-Domain Datasets}

We present the primary results of our method on four standard in-domain BioNER datasets in Table 1. We denote our proposed method, which combines Hybrid Superfiltering with structured JSON generation, as BioSelectTune. The specific instance trained on the 50\% curated positive data subset using the Qwen3-8B base model is referred to as BioSelectTune-8B (50\%). The results are benchmarked against several strong baselines, including general-purpose LLMs (GPT-4), other instruction-tuned models (BioNER-Llama2-7B), and domain-specific pre-trained models (BioMedBERT).

The experimental results, presented in Table 1, powerfully validate our BioSelectTune framework. Our model not only demonstrates highly competitive performance but also achieves new state-of-the-art results, underscoring the superiority of a curated data approach.

A pivotal comparison is against the fully fine-tuned Qwen3-8B-SFT baseline. Our BioSelectTune-8B (50\%) model, despite using only half of the positive training data, outperforms the full-data baseline on three of the four datasets. The most significant improvement is on the NCBI-Disease corpus, where our method achieves a remarkable F1-score of 88.29\%, surpassing the SFT model by a full 1.5 points. This result provides compelling evidence that our Hybrid Superfiltering strategy distills a higher-quality training signal, proving that data quality, not mere quantity, is the key driver of performance.

Furthermore, when benchmarked against the highly specialized BioMedBERT, our method establishes a new state-of-the-art for LLM-based approaches on this task. BioSelectTune-8B (50\%) surpasses the domain-expert model on both the NCBI-Disease and BC5CDR-Disease datasets. This is a critical finding, demonstrating that a general-purpose foundation model, when trained on a meticulously curated dataset via our method, can transcend the performance of models that have undergone extensive and costly pre-training on specialized biomedical corpora. 

It is also instructive to analyze the datasets where BioMedBERT retains a performance advantage, namely BC5CDR-Chemical and BC2GM. We hypothesize that this stems from the fundamental differences between our data-centric fine-tuning and domain-specific pre-training. Entities such as chemical compounds and gene mentions often consist of a vast lexicon of specific proper nouns that rely less on contextual semantics and more on lexical recognition. The exhaustive pre-training of BioMedBERT on biomedical literature likely endows it with a stronger "memory" for this extensive and specific vocabulary. In contrast, our Hybrid Superfiltering method excels at teaching the model to recognize generalizable patterns from complex and ambiguous contexts, making it particularly effective for more descriptively rich entities like diseases. Nevertheless, the fact that our model achieves highly competitive performance across all categories—without the need for costly domain pre-training—powerfully highlights the efficiency and strategic value of our approach. 

\subsubsection{Generalization on Out-of-Domain Datasets}

A critical measure of a model's robustness is its ability to perform well on datasets it has not been trained on. To this end, we evaluated our BioSelectTune-8B model on two completely unseen, out-of-domain datasets: NLM-Gene and NLM-Chem. We benchmarked its performance against the strong, domain-specific BioMedBERT and another instruction-tuned model, BioNER-LLAMA2.

\begin{table*}[htbp]
\centering
\caption{Performance comparison on the NLM-Gene and NLM-Chem datasets (Strict Match). The best F1-scores in each column are highlighted in bold.}
\begin{tabular}{lcccccc}
\toprule
\textbf{Model} & \multicolumn{3}{c}{\textbf{NLM (Gene)}} & \multicolumn{3}{c}{\textbf{NLM (Chem)}} \\
\cmidrule(lr){2-4} \cmidrule(lr){5-7}
& Precision & Recall & Micro-F1 & Precision & Recall & Micro-F1 \\
\midrule
BioMedBERT          & 83.00 & 79.50 & 81.20          & 86.20 & 65.20 & \textbf{74.20} \\
BioNER-LLaMA2       & 88.10 & 77.00 & 82.20          & 85.30 & 59.80 & 70.30          \\
\midrule
BioSelectTune-8B (50\%) & 85.79 & 80.64 & \textbf{83.14} & 85.24 & 64.97 & 73.73          \\
\bottomrule
\end{tabular}
\end{table*}

The results, presented in Table 2, demonstrate that BioSelectTune-8B exhibits excellent generalization capabilities. On the NLM-Gene dataset, our model achieves a new state-of-the-art F1-score of 83.14\%, surpassing both the BioMedBERT baseline and the BioNER-LLAMA2. This indicates that our model learned a more robust and effective representation for gene entity recognition.

Furthermore, on the NLM-Chem dataset, BioSelectTune-8B remains highly competitive with the domain-expert model, achieving an F1-score of 73.73\% and substantially outperforming BioNER-LLAMA2. This strong performance on unseen data suggests that our Hybrid Superfiltering strategy does not lead to overfitting on the in-domain corpora. Instead, it guides the model to learn generalizable features of biomedical entities, underscoring the real-world potential and robustness of our approach.

\subsection{Ablation Studies}

\textbf{Effect of Data Selection Ratio} To determine the optimal quantity of curated data for training, we evaluated the performance of our BioSelectTune-8B model using different ratios of the top-ranked positive samples identified by our filtering strategy. The results are presented in the heatmap in Figure 3.

A key finding from this analysis is that the relationship between the volume of curated data and model performance is non-monotonic. The model achieves its peak performance on three of the four datasets—NCBI-Disease, BC5CDR-Chemical, and BC2GM—when trained on \textbf{50\%} of the curated positive samples. Notably, on NCBI-Disease, the 50\% data subset yields an F1-score of 88.29\%, significantly outperforming both smaller subsets (82.06\% at 25\%) and larger ones, including the full-data baseline (86.78\% at 100\%).

This trend strongly suggests that beyond an optimal point, adding more data that the IFD metric ranked as lower-quality introduces noise that can degrade, rather than improve, final model performance. While the pattern varies slightly across datasets, the overwhelming evidence points to the 50\% mark representing the most effective balance between signal and noise. This result powerfully reinforces the central tenet of our BioSelectTune framework: meticulous data curation is more critical than raw data volume for achieving state-of-the-art performance.

\begin{figure}
\centering
\includegraphics[width=.45\textwidth]{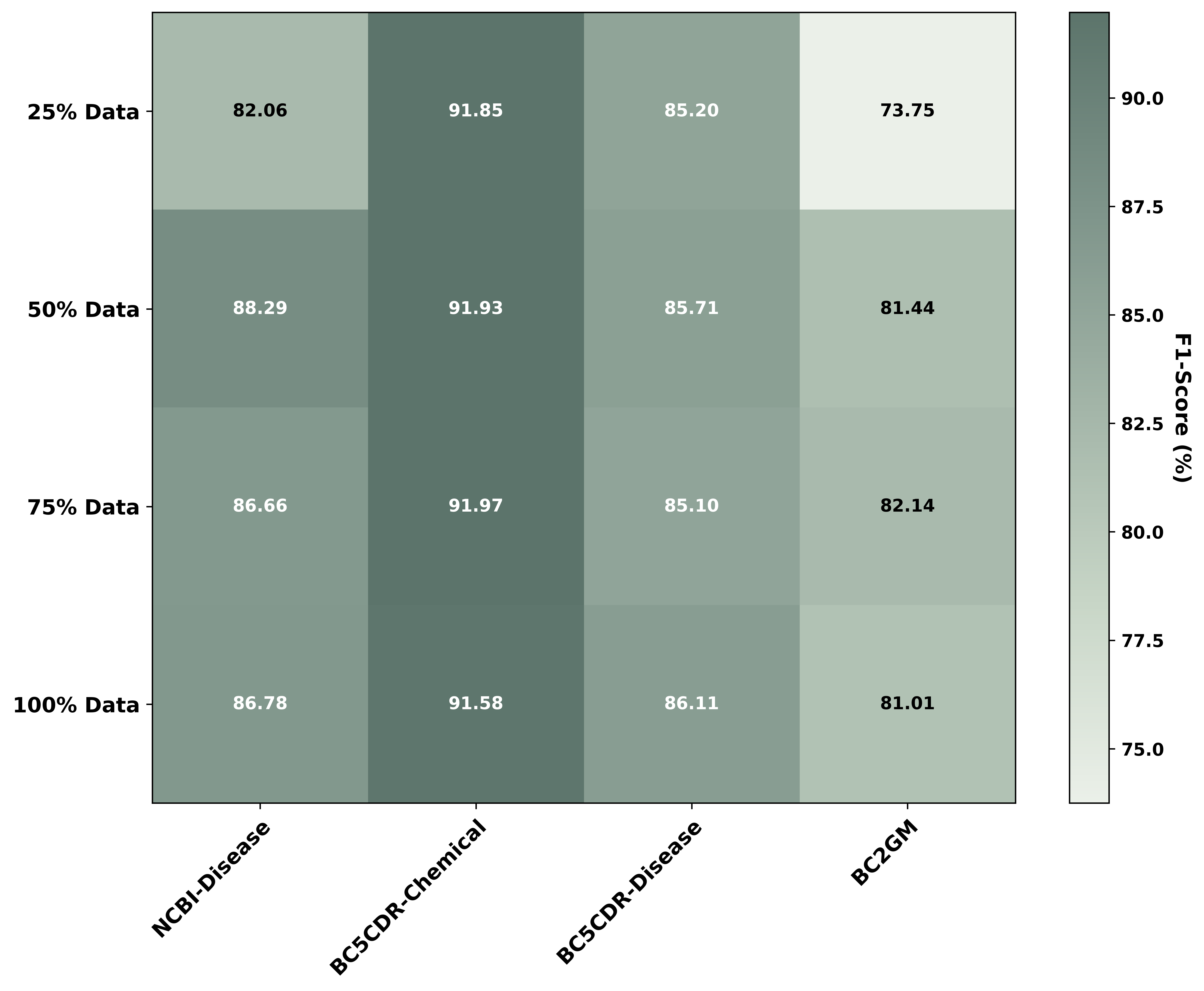}
\caption{Heatmap of F1-scores showing the impact of varying the curated data ratio. The results indicate a clear optimal performance point around the 50\% mark.}
\end{figure}

\textbf{Impact of the Filtering Model.} To validate our choice of a homologous filter, we compared it against using a domain-specific model, BioGPT\cite{luo2022biogpt}, for data curation. We replaced our standard weak filter (Qwen3-0.6B) with BioGPT and used the resulting dataset to fine-tune the Qwen3-8B model, a variant we refer to as BioSelectTune (BioGPT).

The results, presented in Figure 4, are decisive. Our primary method, using the in-family Qwen3-0.6B model, now significantly outperforms the variant filtered by BioGPT on all four datasets. The performance gap is substantial across the board, reaching a remarkable 11.1 F1 points on NCBI-Disease and 9.8 points on BC5CDR-Disease. Even on the BC2GM dataset, where performance was previously similar, our homologous filtering approach now holds a clear advantage of nearly 5 F1 points.

This provides overwhelming evidence for our hypothesis that the architectural and distributional alignment between the filter and target models is far more critical for accurately assessing data difficulty than the filter's specialized domain knowledge. The consistent superiority across all tested corpora validates our choice of a homologous model as the optimal strategy for the BioSelectTune framework.

\begin{figure}
\centering
\includegraphics[width=.45\textwidth]{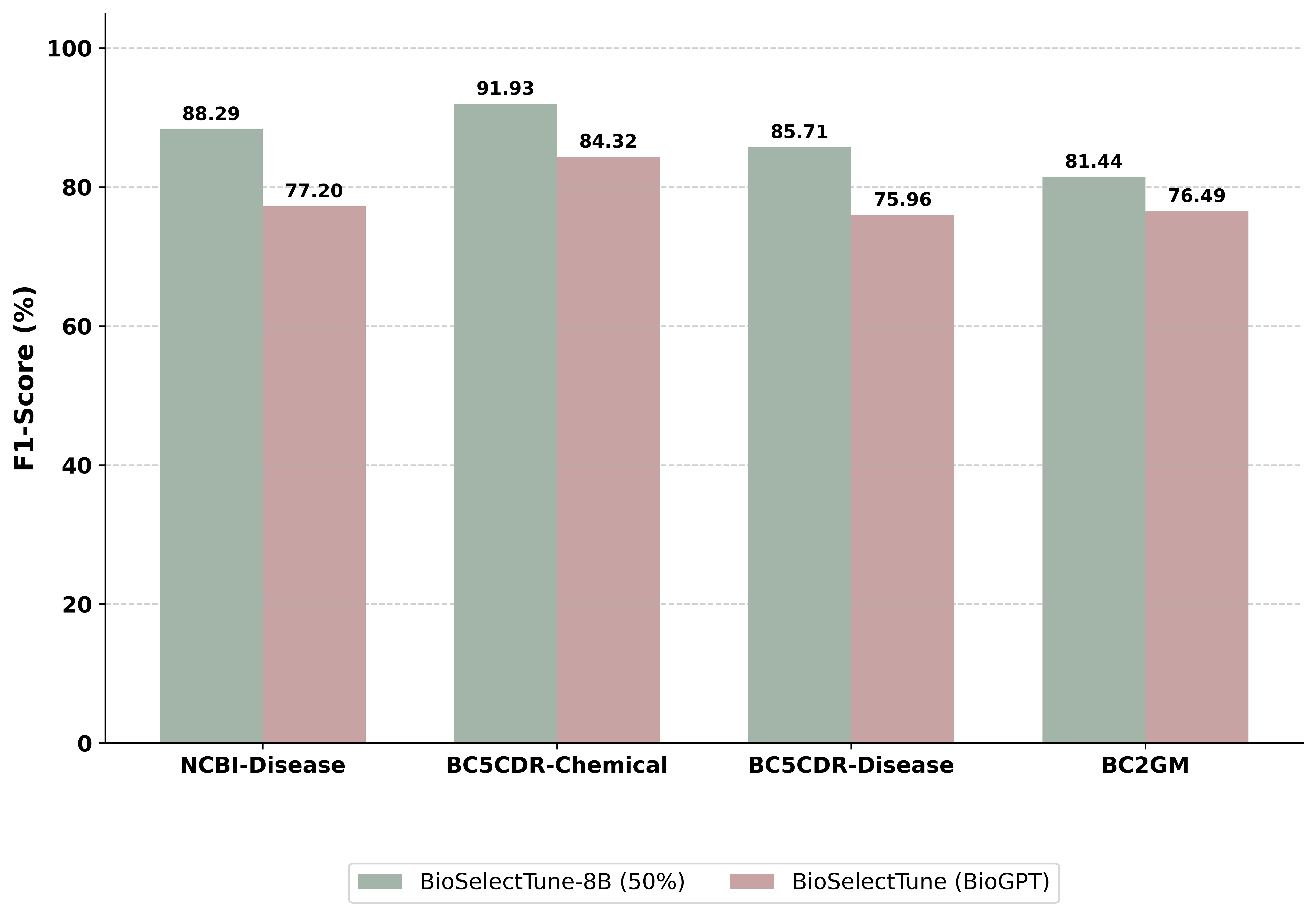}
\caption{Performance comparison between using a homologous weak model (our standard BioSelectTune-8B at 50\%) and a domain-specific weak model (BioSelectTune with BioGPT) for data filtering. The homologous filter demonstrates clear superiority across all datasets.}
\end{figure}

\textbf{Effectiveness on Smaller-Scale Models.} To assess the scalability of our data curation strategy, we applied it to a smaller Qwen3-4B model. The results, presented in Figure 5, compellingly demonstrate that our BioSelectTune framework is highly effective even at a smaller scale. Our method, BioSelectTune-4B (50\%), which uses only half the curated positive data, outperforms the fully fine-tuned Qwen3-4B-SFT baseline on the two largest datasets, NCBI-Disease (84.95\% vs. 84.20\%) and BC5CDR-Chemical (92.07\% vs. 91.71\%). While the full-data baseline retains a marginal advantage on the other two corpora, our findings strongly validate that BioSelectTune can produce smaller models that are not only more efficient to train but can also achieve a superior level of performance.

\begin{figure*}
\centering
\includegraphics[width=.95\textwidth]{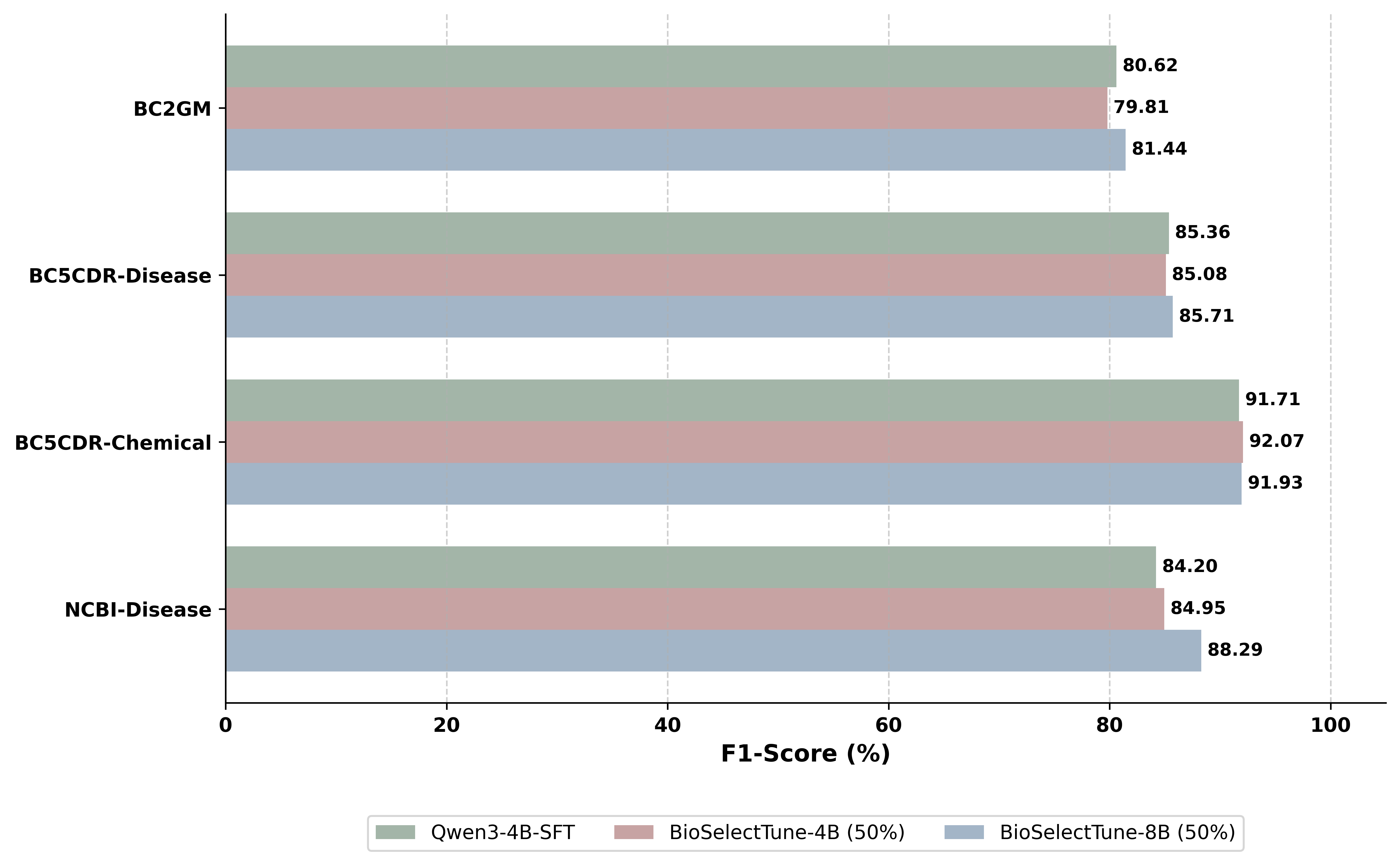}
\caption{Performance comparison of the fully fine-tuned Qwen3-4B-SFT (100\% data) against our BioSelectTune method applied to both the 4B and 8B models (50\% curated positive data). Our strategy enables the smaller 4B model to outperform the full-data baseline on key datasets.}
\end{figure*}


\section{Conclusion}
In conclusion, we introduced BioSelectTune, a data-centric fine-tuning framework that demonstrates the primacy of data quality over quantity in adapting LLMs for BioNER. By combining a structured JSON generation paradigm with our novel Hybrid Superfiltering strategy, our method sets a new state-of-the-art on multiple benchmarks. Notably, our model, trained on only half of the curated positive data, not only surpasses the fully-trained baseline but also outperforms domain-specialized models such as BioMedBERT. The framework's effectiveness, scalability to smaller models, and robust generalization to out-of-domain data provide compelling evidence that our weak-to-strong data curation approach offers an efficient and powerful pathway for unlocking LLMs in specialized scientific domains. Future work will extend this paradigm to more complex biomedical tasks, including relation extraction and document classification.

\printcredits

\section*{Declaration of competing interest}
The authors declare that they have no known competing financial interests or personal relationships that could have appeared to influence the work reported in this paper.

\section*{Acknowledgment}
We thank the anonymous reviewers for their constructive comments.

\section*{Data availability}
Data will be made available on request.

\normalem
\bibliographystyle{unsrt}
\bibliography{ref}
\end{document}